\documentclass[10pt,twocolumn,letterpaper]{article}

\usepackage[pagenumbers]{cvpr} 
\usepackage[accsupp]{axessibility} 

\input{preamble}

\definecolor{cvprblue}{rgb}{0.21,0.49,0.74}
\usepackage[pagebackref,breaklinks,colorlinks,citecolor=cvprblue]{hyperref}

\def\paperID{*****} 
\def\confName{CVPR}
\def\confYear{2024}

\title{Conformal Semantic Image Segmentation: Post-hoc Quantification of Predictive Uncertainty} 

\author{Luca Mossina$^{1, \dagger}$ \quad Joseba Dalmau$^{1}$ \quad L\'{e}o And\'{e}ol$^{2,3}$\\
$^{1}$ IRT Saint Exup\'{e}ry, Toulouse, France\\
$^{2}$ Institut de Math\'{e}matiques de Toulouse, Toulouse, France\\ 
$^{3}$ SNCF, Saint-Denis, France\\
{$^{\dagger}${\tt\small luca.mossina@irt-saintexupery.com}}
}

\begin{document}

\maketitle

\begin{abstract}
We propose a post-hoc, computationally lightweight method to quantify predictive uncertainty in semantic image segmentation.
Our approach uses conformal prediction to generate statistically valid prediction sets that are guaranteed to include the ground-truth segmentation mask at a predefined confidence level. 
We introduce a novel visualization technique of conformalized predictions based on heatmaps, and provide metrics to assess their empirical validity. 
We demonstrate the effectiveness of our approach on well-known benchmark datasets and image segmentation prediction models, and conclude with practical insights. 
\end{abstract}

\section{Introduction}
\label{sec:intro}
\begin{figure}
    \centering
    \includegraphics[width=1.0\columnwidth]{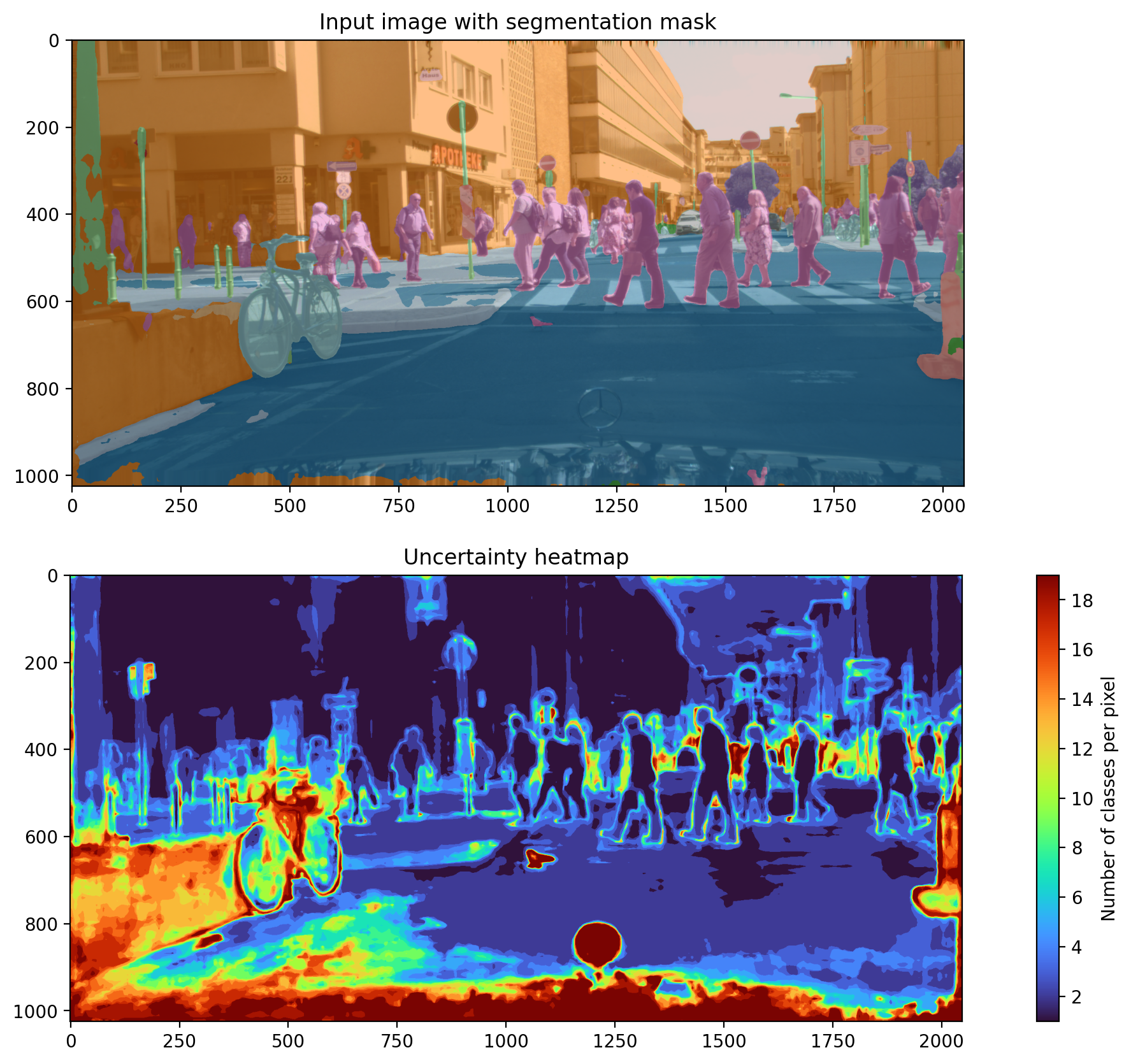}
    \caption{Top: A predicted semantic segmentation mask, overlayed on the input image, for the dataset \emph{Cityscapes} \cite{Cordts_2016_Cityscapes}.
    ~\\
    Bottom: A \vco uncertainty heatmap, for a user-defined risk $\alpha = 0.01$ and a minimum coverage ratio $\tcov$ of $99\%$;
    it is defined in \cref{eq:prediction-set-lac} and statistically valid as in \cref{eq:crc-exp-value-guarantee} of \acf{CRC}:
    every pixel is a prediction set that contains the highest scoring label (top-1) but potentially also the second, third, etc., highest scoring labels.}
    \label{fig:intro-segmask-varisco}
\end{figure}

Despite the success of \ml and \dl models in challenging computer vision tasks such as object detection \cite{Redmon_2016_YOLO, Carion_2020_DETR} or image segmentation \cite{Ronneberger_2015_unet, Mo_2022_Review_sis},  
the complexity of the models makes them akin to black boxes.
It is difficult to define and assess their trustworthiness,
which hinders their adoption in safety-critical industrial applications \cite{Jenn_2020_identifying, mamalet_2021_white, Alecu_2022_can},
and complicates their certification processes \cite{Llorca_2021_trustworthy, EASA_2022_ai}.
In assessing a models trustworthiness, 
the lack of rigorous uncertainty estimates for \ml predictions can be a major drawback, 
notably in the case of \sis \cite{Mo_2022_Review_sis}.
Most segmentation models provide softmax scores (i.e., probability-like scores) for every pixel of an input image; at inference, one builds a segmentation mask by taking the class whose score is the highest, pixel-wise. 
However, softmax scores are known to be overly confident and ill-calibrated
\cite{Guo_2017_calibration, Gupta_2020_calibration};
they tend to yield scores very close to one for the maximum softmax value, sometimes even for ambiguous inputs.
For this reason, softmax values, even if useful for classification purposes, cannot be directly used as measures of uncertainty.

\paragraph{Contributions} 
We introduce a method based on \acf{CP} \cite{Vovk_2005_algorithmic, Papadopoulos_2002_inductive}
to assess the predictive uncertainty of a pre-trained segmentation predictor \f.
Our procedure works with any model \f
(provided that it outputs softmax scores for each pixel)
regardless of its architecture and the distribution of the training data; 
notably, this covers the case of \f being only accessible via an API or being prohibitively expensive to retrain.
Our method quantifies the uncertainty of the predictor \f in the form of \emph{segmentation multi-labeled masks}, that is, segmentation masks that can take multiple labels per pixel. 
Following the conformal algorithm of \cite{Sadinle_2019_lac_Least_Ambiguous},
we build multi-labeled masks as follows: given a \emph{coverage parameter} $\lb \in [0,1]$, the mask $\Clb(X)$ consists, for each pixel in the image, of the labels $c$ having a softmax value higher than $1-\lambda$.
That is, $\forall\text{ pixel } ij$,
\begin{equation}
    \Clb(X)_{ij} = \left\lbrace 
    \text{classes } k \text{ : } \fhat_{ijk} (X)\geq 1-\lb
    \right\rbrace .
    \label{eq:prediction-set-lac}
\end{equation}

\begin{figure*}
    \centering
    \includegraphics[width=\textwidth]{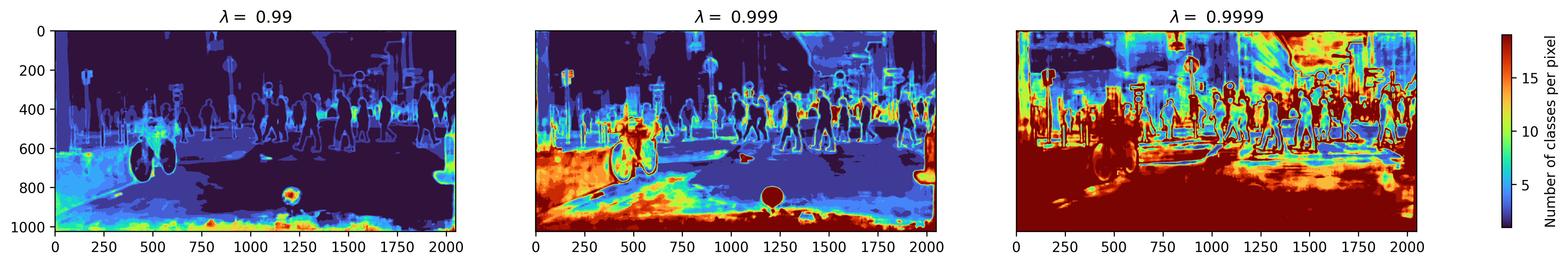}
    \caption{For three (arbitrary) values $\lambda \in \{0.99, 0.999, 0.9999\}$, we apply \cref{eq:cp-lac-threshold} to every pixel and obtain \textit{varisco} heatmaps, {for the dataset \emph{Cityscapes} \cite{Cordts_2016_Cityscapes}}. The \crc algorithm described in \cref{sec:crc-algo} searches for the optimal \lbd  such that, for a given conformalization loss and a risk level $\alpha$, the guarantee in \cref{eq:crc-exp-value-guarantee} is attainable.}
    \label{fig:seg-heatmap}
\end{figure*}

As it can be seen in Figure~\ref{fig:seg-heatmap},
larger values of the coverage parameter \lbd
produce multi-labeled masks with more classes per pixel,
while smaller values of \lbd produce multi-labeled masks
with less classes per pixel.
In order to choose the right value for the coverage parameter \lbd,
the user pre-defines a notion of ``risk'' (or ``error'') 
via a loss function $\ell$ and a maximum tolerable risk $\alpha$.
With the sole requirement of procuring held-out calibration data,
one estimates $\hat{\lambda}$ from the calibration data 
that give rise to the finite-sample, model-agnostic and marginal
guarantee of conformal prediction
\footnote{More precisely, this is the guarantee provided by \crc \cite{Angelopoulos_2024_CRC} which has \cp as a special case.}
\begin{equation}
    \mathbb{E}\big[
        \ell(\C_{\hat{\lambda}}(X_{\text{test}}), Y_{\text{test}})
    \big]
        \leq \alpha.
    \label{eq:crc-exp-value-guarantee}
\end{equation}

The probabilistic guarantee in \cref{eq:crc-exp-value-guarantee} holds under a minimal assumption on the data generation process:
calibration and test data are i.i.d. and statistically independent of the training data.
We also show how these \mulmasks can be visualized by uncertainty \textit{varisco} (visual assessment of risk control) heatmaps, which are computed \textit{post-hoc} with the information of softmax scores.
The code to test our methods can be found at \repo

\section{Background}
\label{sec:background}
\paragraph{Semantic Image Segmentation.}
\acf{SIS} is the task of assigning labels to pixels in an image. 
Let $\mathcal{X}$ be the set of pixel values 
(typically $\mathcal{X}=[0,1]$ for grey-scale images and 
$\mathcal{X}=[0,1]^3$ for color images). 
An image $X$ of $H$ pixels of height and $W$ pixels of width is encoded as the tensor $X = \big\lbrace x_{ij} \in \mathcal{X}\,:\, ij \in \Ihw\big\rbrace$,
where $\Ihw:=\lbrace 1,\dots, H\rbrace\times\lbrace 1,\dots, W\rbrace$
represents the set of indices of the pixels in the image.

Let $\mathcal{L} = \{1, 2, \dots, K\}$ be a set of labels (or ``classes''); each pixel $x_{ij}$ is associated to one label $y_{ij} \in \mathcal{L}$.
The set $Y = \{y_{ij} \in \mathcal{L}\,:\, ij\in\Ihw  \}$ is commonly referred to as the \emph{segmentation mask} of the image $X$ (see Fig.~\ref{fig:intro-segmask-varisco}), and the goal of the \sis task is to infer the segmentation mask $Y$ given the image $X$. 
This is typically done by training a predictor \f that outputs softmax values for each pixel.

\paragraph{Conformal Prediction.}
\acf{CP} \cite{Vovk_2005_algorithmic, Angelopoulos_2021_Gentle} is a family of uncertainty quantification techniques
that provide model-agnostic, finite-sample guarantees on 
the predictions of \ml models.
The most common \cp technique, split \cp \cite{Papadopoulos_2002_inductive}, 
is applied post-hoc on a trained model \f. 
It requires a calibration dataset $\{ (X_1, Y_1),...,(X_n Y_n)\}$ independent of the 
training data, and an acceptable error rate $\alpha \in (0,1)$ set by the user.
Split \cp uses nonconformity scores (to be understood as a form of measure of
prediction error) computed on the calibration dataset in order to build a prediction set $\C_{\alpha}(X_{n+1})$ for a new test sample $X_{n+1}$.
The guarantee achieved by using split \cp is
\begin{equation}
    P \big( Y_{n+1} \in \C_{\alpha}(X_{n+1}) \big) \geq 1-\alpha.
    \label{eq:split-cp}
\end{equation}
The only assumption is that the calibration and test data form an exchangeable sequence (a condition weaker than forming an i.i.d. sequence)
and that they are independent of the training data. 
The main limitation of \cp is that the guarantee in \cref{eq:split-cp} is marginal, i.e. it holds on average over both the choice of the calibration
dataset and the test sample.

\paragraph{Conformal Risk Control.}
In many applications, errors of different nature 
may have a different severity,  a false negative
vs a false positive in a tumor detection application.
The notion of severity of an error
can be captured via a risk or an error function.
\ac{CRC} \cite{Angelopoulos_2024_CRC} generalizes the ideas of Conformal Prediction
to this setting: prediction sets are guaranteed to 
keep the expected risk below a user pre-defined level $\alpha$.
We show how to adapt this approach to \sis in \Cref{sec:crc-segmentation}.
Note that when using binary losses, the guarantee of \crc is the same as that of \cp in \cref{eq:split-cp}.

\section{Related works}
\label{sec:related-work}

State-of-the-art \ml predictors, based on deep learning, are so complex that they are commonly approached as black boxes: 
the users provide some input data (an image) and they retrieve a prediction. 
How accurate are these models? 
The study of this subject is known as \uq and is a key element towards building trustworthiness in systems powered by ML models \cite{Mattioli_2023_towards, Mattioli_2023_overview}.

Uncertainty is commonly conceptualized \cite{Huellermeier_2021_Aleatoric, Lahoti_2022_Responsible} as having an \emph{aleatoric} and \emph{epistemic} component.
{Aleatoric} uncertainty is inherent to the modeled phenomenon and non reducible. 
{Epistemic} uncertainty, on the contrary, stems from the fact that the models we use do not capture the phenomenon being modeled faithfully enough, 
and can usually be reduced by taking into account new observations
or by enriching the model family being used.
\cp provides an estimation of the global uncertainty in the model's predictions, since it is {post-hoc} and with minimal hypotheses.

\subsection{Uncertainty quantification methods for semantic image segmentation}
\label{sec:related-work-uq-seg-non-cp}
Some approaches to UQ leverage model architectures that provide not only a point prediction but also the associated uncertainty, notable examples include Bayesian neural networks, networks based on Monte Carlo techniques and Deep Ensembles;
others fit auxiliary models.
\cite{Krygier_2021_quantifying} use Monte Carlo dropout neural network as well as Bayesian neural networks; 
\cite{Mukhoti_2023_deep_uncertainty} combine Gaussian Discriminant Analysis to estimate 
the density of the feature-space with the entropy measure of the softmax predictions in order 
to disentangle epistemic and aleatoric uncertainty and estimate each of them separately;
in a similar vein, \cite{kendall_2017_SegNet, Kendall_2017_uncertainties} train a bayesian neural network in order to estimate point-predictions, 
{aleatoric} uncertainty and {epistemic} uncertainty simultaneously.
\cite{S_2021_uncertainty} work on reducing the uncertainty of \sis models
within the framework of model adaptation, that is, when domain adaptation is to be performed in the absence of source data.
We also refer to \cite{Corbiere_2019_failure_prediction}, who work on failure prediction, a subject related to \uq.
They train an auxiliary model to score the confidence of a prediction.
They detect when a predictor has made a wrong prediction and assess this via a confidence measure. 
With respect to these approaches, for our contribution we restrict the scope to post-hoc methods: we suppose to be given a pretrained predictor that we cannot modify and that the training data are not accessible, while providing a theoretical guarantee on the \uq.

\subsubsection{Calibration of image segmentation}
A well-established approach to \uq is that of the \textit{calibration} of ML models that output (pseudo) probabilities for the labels, where either auxiliary models or empirical adjustments to models are employed.
It is known \cite{Guo_2017_calibration} that these scores do not admit a valid probabilistic interpretation, notably for deep-learning models based on the minimization of cross-entropy.
\cite{Guo_2017_calibration}, among others, brought this concept to the attention of the ML community, studying how calibration methods such as Platt scaling \cite{Platt_1999_probabilistic} may yield a probabilistically valid interpretation.
This notion of uncertainty is applicable \cite{Czolbe_2021_is_segmentation_useful, Rousseau_2021_posttraining, Judge_2022_crisp_uncertainty, LourencoSilva_2022_Soft_labels, Yang_2023_uncertainty_reward}
to semantic image segmentation, where each pixel embeds a multiclass classification problem.
\cite{Cygert_2021_look, Rousseau_2021_posttraining}, for instance, give some empirical results on the effect of several methods on calibration errors.
Similary, \cite{Bohdal_2023_label} have used calibration methods to address the issue of domain shift.
Recently, \cite{Wang_2023_calibrating} 
proposed selective scaling as a means to calibrate the segmentation softmax scores.
{These methods could be used as a complement to \cp \cite{Xi_2024_calibration_conformal}, at the cost of training an additional model.}
Finally, some literature \cite{Gupta_2020_calibration, Gupta_2022_toplabel} does provide theoretically-founded, distribution-free algorithms for calibration and study their connection to \cp \cite{Gupta_2020_cp_calib}.
Our work is based on Conformal Prediction, which is not a calibration method, but rather a different technique of \uq. 
As such, it is complementary to calibration, and can be used both with a model's original output as well as with an output that has been previously calibrated.

\subsection{Applications of CP to segmentation}
\label{sec:related-work-cp-seg}
\cite{Angelopoulos_2024_CRC} use their \crc to control the false negative rate in tumor segmentation.
Also based on risk control, albeit using different mathematical frameworks, the contributions of \cite{Park_2020_PAC, Bates_2021_RCPS, Angelopoulos_2021_LTT} extend the concept of tolerance regions to \ml problems.
They offer stronger guarantees at the cost of inferior sample efficiency.
We refer to \cite{Angelopoulos_2021_Gentle} for an introduction.
Of these, \cite{Bates_2021_RCPS} apply their methods to binary segmentation of medical images. 
 
As for existing work using \cp based on nonconformity scores,
\cite{wieslander2021tissue} apply \cp to medical imaging, building pixel-wise confidence scores based on nonconformity scores and p-values \cite{Vovk_2005_algorithmic}.
\cite{Teng_2023_predictive} compute the nonconformity scores in the feature space and present an application to image segmentation.
For the case of \cp in imaging, we also point out to the literature on image-to-image regression (image reconstruction) \cite{Angelopoulos_2022_Image_to_Image_Reg, Kutiel_2022_whats, Teneggi_2023_k_RCPS, Belhasin_2023_volumeoriented} which builds intervals for each output pixel.
Previous work using \crc for semantic image segmentation focuses on the binary segmentation case.
To the best of our knowledge, our work is the first that addresses the multi-class segmentation task with the theoretical guarantee of conformal risk control.

\section{Conformal Semantic Segmentation}
\label{sec:conformal-semantic-segmentation}
The goal of conformal semantic segmentation is to produce prediction sets that remain below a user-specified risk.
The prediction sets are then used to assess the behaviour of the underlying predictor \f together with the problem data.

\subsection{Multi-labeled masks.}

A prediction set will take the form of a \emph{\mulmask}, that is, a tensor 
\begin{equation} 
    Z = \big\lbrace z_{ijk} \,:\, ij \in\Ihw, k\in\mathcal{L} \big\rbrace,
\label{eq:notation-multi-mask}
\end{equation}
where $(z_{ijk})_{k=1}^K\in\lbrace 0,1\rbrace^K$ 
encodes the subset of labels corresponding to the pixel $ij$;
Note that this tensor has as many channels as the number of classes, where each channel is a binary segmentation mask (class $k$ vs others).

For a multiclass segmentation mask $Y$, its one-hot encoding $h(Y)$
is a particular instance of a \mulmask: every pixel has exactly one channel (out of $K$) with value one.
We say that a \mulmask $Z$ contains a \mulmask $Z'$ and we write $Z \geq Z'$,
if $z_{ij} \geq z'_{ij}$ for each pixel $(i,j)$.

\subsubsection{Nested \mulmasks.}
\label{sec:nested-multimasks}
Let \smash{\f} be any semantic segmentation predictor that produces pixel-wise softmax scores,
that is, for an image $X$, we have
\begin{equation}
    \widehat{f}(X) := \big\lbrace\widehat{f}_{ijk}(X)\,:\, ij \in\Ihw, k \in \mathcal{L} \big\rbrace,
\end{equation}

\noindent with $\smash{\fhat_{ijk}(X)}\in [0,1]$ and $\sum_{k = 1}^{K} \fhat_{ijk}(X) = 1$. 
Our baseline conformal segmentation method builds prediction \mulmasks based on the point-predictor \smash{\f}, 
via the \lac \cite{Sadinle_2019_lac_Least_Ambiguous}.
Given $\lambda \in [0,1]$ and a probability $p\in [0,1]$,
we define the thresholding $T_\lambda(p)$ by setting:
\begin{equation}
    T_\lambda(p)=\begin{cases}
        \quad 1 &\quad \text{if } p\geq 1-\lambda,\\
        \quad 0 &\quad \text{otherwise}.
    \end{cases}
\end{equation}

The \lac mapping on the whole image $X$ is defined by applying the mapping $T_\lambda$ to the tensor $\fhat(X)$
\begin{equation}
    \Clac_{\lb}(X) = \big\lbrace
    T_\lambda\big(\widehat{f}_{ijk}(X)\big)
    \,:\, ij \in \Ihw, k\in\mathcal{L}\big\rbrace.
    \label{eq:cp-lac-threshold}
\end{equation}

The \mulmasks generated via the LAC procedure are nested,
i.e.
\begin{equation}
    \lambda \leq \lambda' \quad\Rightarrow\quad \Clac_\lambda \leq \Clac_{\lambda'}.
\end{equation}

Note that following \cref{eq:crc-lb-hat}, for high values of $\alpha$, one can get a valid $\lbhat$ that can produce some empty pixels (i.e. there is no class with a high-enough score) 
when plugged into \cref{eq:cp-lac-threshold}.
To prevent this,
we always include the most highly scored class in the \mulmask (or ``top-1'' class).

\subsection{Conformal Risk Control for \mulmask} 
\label{sec:crc-segmentation}
Instead of working with loss functions that compare ground-truth values to point-predictions,
\acf{CRC} considers loss functions that compare ground-truth values to set-predictions.
For the particular application of semantic segmentation, 
we consider a loss function $\ell(Z, Y)$ taking as arguments a \mulmask $Z$ and a one-hot
encoded mask $Y$. 
We assume that $\ell$ takes values in the bounded interval $(-\infty, B]$ for some $B\in\mathbb{R}$, 
and that it is non-increasing in $Z$:
\begin{equation}
\label{eq:monotonous_ell}
    \forall\, Y,\ \forall\, Z \leq Z' \quad \Rightarrow \quad 
    \ell(Z, Y) \geq \ell(Z', Y),
\end{equation}
i.e. larger \mulmasks produce smaller loss values.
This assumption, together with the nestedness of the \lac masks imply that the mapping 
$\lambda \mapsto \ell\big(\Clb^{LAC}(X),Y\big)$
is non-increasing in \lbd.
The loss $\ell$ is not used as a training loss, 
that is, applying \crc does not affect the pre-trained model nor the predictive algorithm directly.
The loss $\ell$ rather allows the user to \textit{encode the notion of error} \cite{Angelopoulos_2024_CRC} in their predictions.
In order to be clear, the loss $\ell$ will be referred to as \textit{conformalization loss}.

Let us consider a sequence $(X_i,Y_i)_{i=1}^{n+1}$ of images with their corresponding ground-truth segmentation masks.
The first $n$ examples constitute our calibration set $\Dcal$ and the example $n+1$ is taken to be the test example.
We denote $L_i(\lambda):=\ell\big( \C_\lambda(X_i), Y_i\big)$
the loss on the $i$-th example, 
one can then compute the \emph{empirical risk} of the prediction sets on calibration data $\Dcal$ as
\begin{equation}
    \Rhat_n(\lb) = \frac{1}{n}\sum_{i=1}^{n} L_i(\lb).
    \label{eq:emp_risk}
\end{equation}
The purpose of the calibration set is to estimate the right value 
$\lbhat$ with the guarantee that the risk will remain below 
the maximum tolerated risk level. 
Given a maximum tolerated risk level $\alpha\in\mathbb{R}$, we define
\begin{equation} 
    \lbhat := \inf\left\{ \lb \in [0,1] : \frac{n}{n+1} \Rhat_{n}(\lb) + \frac{B}{n+1} \leq \alpha  \right\}
\label{eq:crc-lb-hat}
\end{equation}

\begin{theorem}[Theorem 1 in \cite{Angelopoulos_2024_CRC}.]
\label{th:crc-guarantee} 
Assume that the $L_i(\lb)$ are non-increasing, right-continuous and bounded by $B<+\infty$.
Assume that there exists 
$\lbmax \in [0,1]$ 
such that $L_i(\lbmax) \leq \alpha$.
Assume further that $L_1(\lb),\dots,L_{n+1}(\lb)$ form an exchangeable sequence.
Let $\smash{\lbhat}$ be computed as in Equation ~\eqref{eq:crc-lb-hat}.
Then it holds that
\begin{equation}
    \mathbb{E} \left[ L_{n+1}(\lbhat) \right] \leq \alpha.
    \label{eq:crc-loss-guarantee}
\end{equation}
\end{theorem}

\paragraph{Computing the optimal $\lbhat$.}
\cite{Angelopoulos_2024_CRC} do not provide an explicit optimization algorithm to find the optimal parameter in \cref{eq:crc-lb-hat}.
Because of the hypothesis of monotonicity of $L_i(\lb)$ with respect to $\lb$, this can be achieved, for instance, running a dichotomic search over the parameter \lbd, up to any user-defined error  $\epsilon$.

\subsubsection{Conformalization algorithm}
\label{sec:crc-algo}
When we say we ``conformalize'' a \ml predictor, we mean computing the losses and the optimal $\lbhat$ on some calibration data. More specifically:

\begin{algorithm}
\caption{Conformalization of Semantic Image Segmentation, setup and estimation of $\lbhat$.}
\label{alg:crc-lb-opt}
    \KwData{Predictor $\fhat \in [0,1]^{K \times H \times W}$.
    Prediction set parametrization $\Clb(\cdot)$.
    }
    \KwResult{$\lbhat$}
    Collect calibration data $\Dcal = (X_i, Y_i)_{i=1}^{n}$ from the same distribution as the test data;
    
    Choose a conformalization loss $\ell(\Clb(X), Y) \in [0,1]$ (see \cref{sec:loss-selection});
         
    Set an acceptable risk level $\alpha \in (0,1)$;
    
    Compute $\lbhat$ as in Eq.~\eqref{eq:crc-lb-hat}: since the empirical risk is monotonic w.r.t. \lbd,  dichotomic search is a fast option;
\end{algorithm}

\noindent
Note that to ensure the statistical validity, one must pick a value $\alpha$ before observing the calibration data: like in statistical hypotheses testing, one cannot adjust their significance level $\alpha$ after computing the p-values.
In practice, one could use two calibration datasets, the first to explore \cp on the use case and the second reserved to the estimation of the $\lbhat$ to be deployed in production.

We say we have a ``conformalized prediction'' when we build the prediction set with the $\lbhat$ as computed above, applying \cref{eq:cp-lac-threshold}.

\begin{algorithm}
\caption{Conformalization of Semantic Image Segmentation, {inference}.}
\label{alg:crc-inference}
    \KwData{Input image $X \in [0,1]^{3\times H \times W}$, Predictor $\fhat \in [0,1]^{K \times H \times W}$.
    Prediction set parametrization $\Clb(\cdot)$}
    \KwResult{$Z = \Clac(X)$}

    Compute $\Clac(X)$: apply Eq.~\eqref{eq:cp-lac-threshold} to $X$.
\end{algorithm}

\subsection{Choosing the loss function.}
\label{sec:loss-selection}
\begin{figure*}
    \centering
    \includegraphics[width=\textwidth]{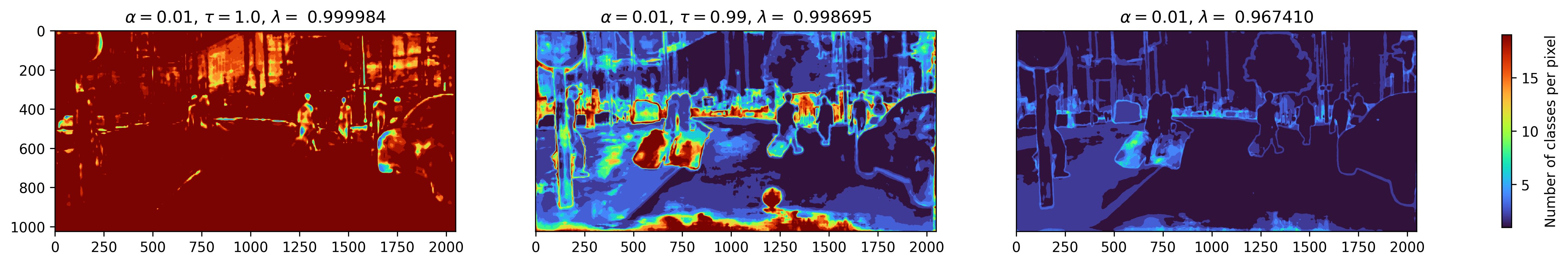}
    \caption{For the same risk level $\alpha=0.01$, different losses yield different heatmaps: (left) binary loss $\ell_{\text{bin}}$, (center) binary loss with threshold $\ell_{\tcov}$, (right) miscoverage loss $\ell$. 
    If the notion of risk is too restrictive, the prediction set will be theoretically valid but not very informative. 
    In this example, the figure on the left (binary loss, $\tcov = 1.0$) has most of the pixels of color red, indicating that $K$ (out of $K$) classes are in the prediction set. Dataset: \textit{Cityscapes} \cite{Cordts_2016_Cityscapes}.}
    \label{fig:compare-losses-and-alphas}
\end{figure*}

In \cref{fig:compare-losses-and-alphas} we show how, for the same risk level $\alpha=0.1$, different losses generate different \vco heatmaps.
These losses encode different notions of error: from the left to right-hand side, we see a shift from stricter to less strict.
When implementing a \cp method, the users need to choose a conformalization loss $\ell$ suitable to their problem. 
For the guarantee of \crc to hold, 
one needs to ensure that the losses $L_i(\lb)$ are non-increasing with respect to \lbd;
since the \lac procedure in Eq.~\eqref{eq:cp-lac-threshold} produces nested prediction sets \cite{Gupta_2022_nested}, it is enough to ensure that $\ell(Z,Y)$ is non-decreasing with respect to the first argument $Z$.
In this section we give three examples of natural choices for losses that respect this condition.

The first is a \textbf{binary loss}, which yields a guarantee equivalent to that of \cp based on nonconformity scores, whose underlying loss would be $\ell(\C(X), Y) = \mathbbm{1}\{Y \not \in \C_{\alpha}(X)\}$.
It takes value one whenever the prediction set does not contain the true value Y. 
We write as
\begin{equation}
    \ell_{\text{bin}}(Z, Y) = 
    \begin{cases}
        0 & \text{if}\quad Z \geq Y\\ 
        1 & \text{otherwise}.
    \end{cases}
    \label{eq:bin_loss}
\end{equation}
In the case of conformalized \sis,
that happens when the \mulmask does not cover every pixel in the image.
Empirically, the conformalization of segmentation produces very small values $\lbhat$ that result in \mulmasks corresponding to (almost) the whole target space $\Y = \{1, 1, \dots, 1 \}^{K \times H \times W}$ for each inference (\eg left-hand side in \cref{fig:compare-losses-and-alphas}).

One can however set an acceptable trade-off in coverage, with a \textit{minimum coverage ratio} $\tcov$: the user specifies a priori the minimal proportion of pixels that need to be covered for a prediction to be considered successful.
We thus define the \textbf{binary loss with threshold} as
\begin{equation}
    \ell_{\tau}(Z, Y) = 
    \begin{cases}
        1 & \text{if}\quad \frac{\sum_{ijk} Z_{ijk}Y_{ijk}}{\sum_{ijk}Y_{ijk}} < \tau,\\ 
        0 & \text{otherwise}.
    \end{cases}
    \label{eq:bin_loss_threshold}
\end{equation}

\noindent where $\tcov$ is the minimum acceptable coverage ratio. 
For $\tcov = 1$ we recover the binary loss in Eq.~\eqref{eq:bin_loss}.

Binary losses constitute a strict criterion: 
during conformalization, for $\tcov = 90\%$, an empirical coverage of $89.9\%$ will be considered a failure.
A less strict notion of error is given by directly controlling the coverage via the \textbf{miscoverage loss}
\begin{equation}
    \ell(Z, Y)
        = 1 - \frac{\sum_{ijk} Z_{ijk}Y_{ijk} }{ \sum_{ijk}Y_{ijk} } 
    \label{eq:loss_miscoverage}
\end{equation}
The miscoverage loss is directly related to the concept of accuracy \cite{hu2016recalling,zhang2022resnest,chen2022few} and can be easily extended to follow the balanced accuracy
known in the medical literature \cite{anthimopoulos2018semantic, attia2017surgical} or even a weighted version (\eg lower importance to background pixels), inspired for instance by \cite{can2021semantic,ling2023mtanet}.
\footnote{
Example of weighted miscoverage loss: 
\begin{equation}
    \ell_{w}(Z, Y)
        = 1 - \frac{1}{\sum_{k} w_k} \sum_{k}w_k \frac{\sum_{ij} Z_{ijk}Y_{ijk} }{ \sum_{ij}Y_{ijk} } 
    \label{eq:weighted-loss_miscoverage}
\end{equation}
}

\subsection{{Varisco} heatmaps}

In \cp, the size a prediction set (\eg a prediction interval) is taken as a signal of uncertainty: for a risk set by the user, it corresponds to the ``typical'' error measured in the calibration dataset.
In our case of image segmentation, we look at every pixel in the output: how many classes there are, whose softmax score is above the threshold $\lbhat$.
If we count the labels in each pixel and normalize by $K$, 
we can generate an image which has, for each pixel, a scalar value in $[0,1]$.
Mapping these scalars to a gradient of colors we obtain a \textbf{heatmap} corresponding to the underlying prediction \mulmask.
 
In Figure~\ref{fig:seg-heatmap} are three examples: for the same predicted softmax, we apply thresholds $\lb \in \{0.99, 0.999, 0.9999\}$ and obtain three different heatmaps.
When \lbd is computed with a \crc procedure on calibration data,
these heatmaps provide a qualitative visualization of the model's uncertainty
obtained from the risk control procedure,
hence the name \emph{varisco} (visual assessment of risk control).
Furthermore, for a better visualization in datasets with many classes (\eg \emph{LoveDA}, see Section~\ref{sec:experiments}), scaling the class count in every pixel by the maximum count observed in the \mulmask (often $\ll K$) is also helpful.

The use of heatmaps is not new in semantic image segmentation, and one can find recent examples in \cite{Chan_2021_segmentmeifyoucan, Chan_2021_entropymaximization}, where they are used for Out-of-Distribution (OOD) detection or in some of the \uq literature cited in Sections~\ref{sec:related-work-uq-seg-non-cp} and ~\ref{sec:related-work-cp-seg}.
To the best of our knowledge, however, this is the first time that this kind of visualization based on prediction sets is mentioned in the context of \uq and \cp, with their underlying theoretical guarantee.

\subsubsection{Characteristics of heatmaps}
Our \vco heatmaps are monotone in the parameter $\lambda$: 
as $\lambda$ grows, the set of pixels for each class is non-decreasing in size.
Note that the heatmaps contain information about the aleatoric and epistemic uncertainty.
As a general rule of thumb, for semantic segmentation tasks the aleatoric uncertainty should 
be maximal around the edges of the ground-truth figures, so that a heatmap with warm regions 
away of the contours should warn the user that the epistemic uncertainty of the model is high,
and better models might be available for the data at hand.

The parameter \lbd encapsulates a notion of conservativeness, 
i.e. the higher the parameter \lbd, the more activation we will get in our \mulmask,
and therefore in our heatmap.
The calibration of \lbd corresponds to the user setting an acceptable risk $\alpha$ and finding the least conservative \lbd such that their need is met.
Note that for an arbitrary $\lambda$, the associated heatmap provides little information
about the epistemic uncertainty of the model \f, meaning that,
given two different point predictors $\widehat{f}_1$ and $\widehat{f}_2$,
plotting the heatmaps $H_1(\lambda)$ for the first model and $H_2(\lambda)$ will give us 
no information about which of the two models performs best.
This is because the heatmaps $H_i(\lambda)$ carry no information about the errors of the models,
but rather about the entropy of the softmax-es in each model.
However, given a pre-set risk level $\alpha$, 
once the appropriately values $\widehat{\lambda}_1$ and $\widehat{\lambda}_2$ are chosen
through the CRC calibration procedure,
we can safely compare the heatmaps $H_1(\widehat{\lambda}_1)$ and $H_2(\widehat{\lambda}_2)$,
because both heatmaps guarantee the same risk level for both models. 
A warmer heatmap for the model $\widehat{f}_1$ means that for the chosen risk level $\alpha$,
the model $\widehat{f}_1$ carries more epistemic uncertainty than the model $\widehat{f}_2$.

\section{Metrics and UQ diagnostics}\label{sec:metrics-diagnostics}
To the best of our knowledge, ours is the first paper that uses prediction sets via \mulmasks to provably quantify the uncertainty in multiclass semantic segmentation; 
it is not possible to compare our results directly to existing approaches (see Section~\ref{sec:related-work}), as they are \textit{essentially} different.
However, as it is common in the \cp literature, one can test different nonconformity scores or, as in our case, the coupling of set parametrization (\eg $\Clac$) and conformalization loss.
Throughout the paper we restrict our exposition to the nested-set parametrization derived from the algorithm of \cite{Sadinle_2019_lac_Least_Ambiguous}, which we refer to as \lac.
Although out-of-scope for this paper, 
our code repository (see \cref{sec:intro}) contains some examples using the \ac{APS} algorithm of \cite{Romano_2020_APS}, which employs a threshold on the sum of the softmax scores sorted in decreasing order. 

To assess experimentally the validity of \crc, we compute the \textbf{empirical risk} as in \cref{eq:emp_risk} on the test data.
In \cp, a standard metric of the efficacy of the method 
is the average size of the prediction sets on the test data.
For classification, this boils down to counting the number of classes in the prediction set, averaged over the test set.
With semantic segmentation, this can be thought of as computing the average number of ``activated'' classes (\ie, whose softmax is above the threshold $\lbhat$)
over all pixels in the input image. 
For a one-hot-encoded \mulmasks $Z$ defined as in \cref{eq:notation-multi-mask}, 
the \textit{prediction set size} of a pixel $(i,j)$ is $\sum_{k=1}^{K}z_{kij}$.
Extending this to the whole image and normalizing by the number of valid pixels $n_{\text{pixels}}$ (\eg excluding void pixels, artefacts, etc. common in computer vision), we have, for one \mulmask $Z$, the \textbf{activation ratio}
\begin{align}
    \text{AR}(Z) =& \frac{1}{n_{\text{pixels}}} \sum_{i,j,k} z_{ijk},\\
    &\text{ for all pixels } x_{ij} \text{ that were labeled in } X. \nonumber
    \label{eq:metrics-multimask-activation}
\end{align}

\section{Experiments}\label{sec:experiments}
\begin{figure*}[h]
    \centering
    \includegraphics[width=1.0\textwidth]{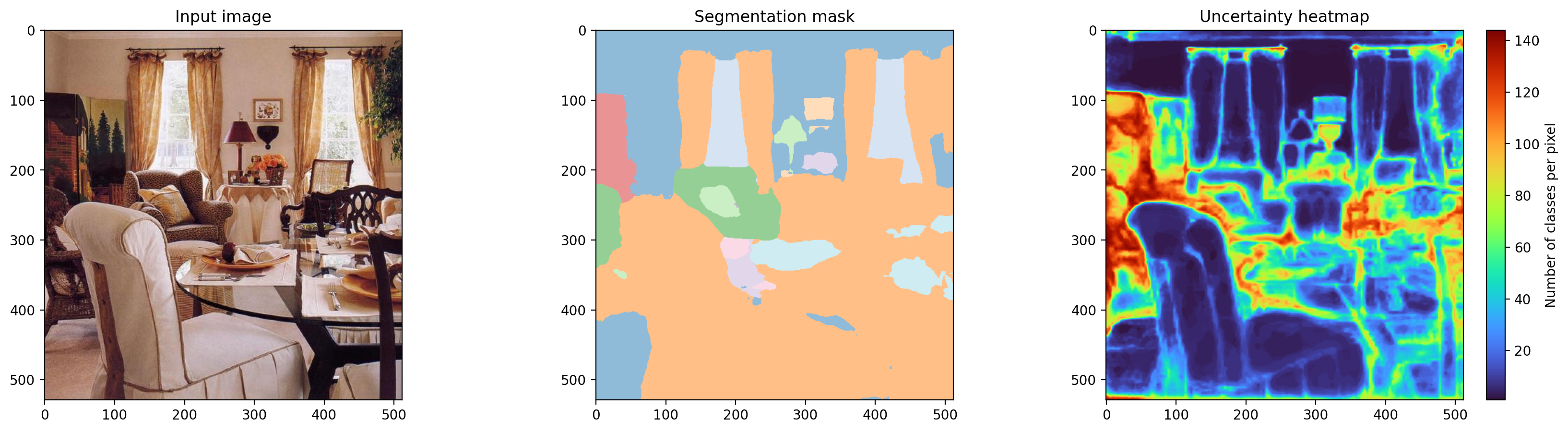}
    \caption{Visualization of a \vco heatmaps (miscoverage loss, $\alpha = 0.01$) for the \textit{ADE20K} dataset \cite{Zhou_2017_Scene, Zhou_2019_Semantic}: (left) input image, (center) predicted segmentation mask, (right) \vco heatmap.}
    \label{fig:visu-}
\end{figure*}

\begin{figure*}[h]
    \centering
    \includegraphics[width=1.0\textwidth]{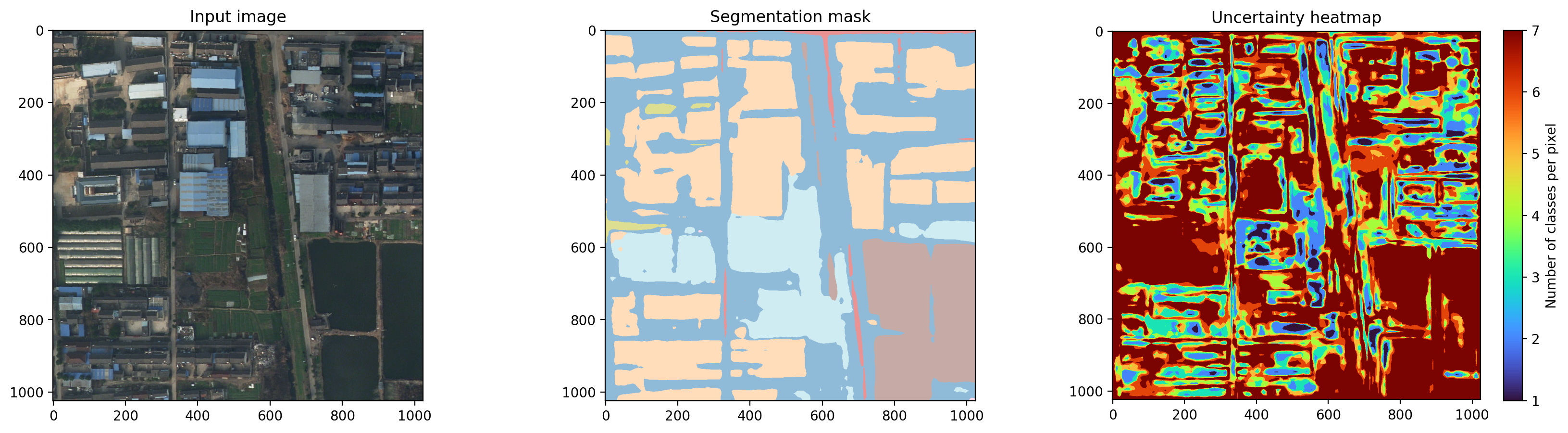}
    \caption{Visualization of a \vco heatmaps (miscoverage loss, $\alpha = 0.01$) for the \textit{LoveDA} dataset \cite{Wang_2021_LoveDA, Wang_2021_LoveDA_dataset}: (left) input image, (center) predicted segmentation mask, (right) \vco heatmap.}
    \label{fig:visu-conformal-maps}
\end{figure*}

Since \cp is model-agnostic, we test architectures and datasets of varying size and complexity, knowing that regardless of the predictor chosen, our method will be statistically valid.
For all the models tested, we used the PyTorch \cite{Paszke_PyTorch_An_Imperative_2019} implementation provided by the open-source Python library \textit{MMSegmentation} \cite{mmseg_2020}, which includes code to run inferences as well as pretrained weights for many datasets.

We run our experiments on \emph{Cityscapes} \cite{Cordts_2016_Cityscapes} (19 classes, automotive vision), 
\emph{ADE20K} \cite{Zhou_2017_Scene, Zhou_2019_Semantic} (150 generic common classes)
and \emph{LoveDA} \cite{Wang_2021_LoveDA, Wang_2021_LoveDA_dataset} (aerial images, 7 classes).
As for the architectures of the neural networks, we selected the best performing ones within our computational budget: PSPNet \cite{Zhao_2017_pspnet} for \textit{Cityscapes} and \textit{LoveDA}, SegFormer for \textit{ADE20K} \cite{Xie_2021_segformer}.

For conformalization, we split the validation data into two partitions, one for calibration and one for testing.
We tested both the binary loss with threshold of Eq.~\eqref{eq:bin_loss_threshold} and the miscoverage loss of Eq.~\eqref{eq:loss_miscoverage}.
For the risk $\alpha$, we tested values that would have made sense in the real world: for \textit{Cityscapes}, our pretrained model has a very good performance (\eg mIoU) and the user can aim for small risks.
For the other cases, such small $\alpha$ could yield hardly informative prediction sets.
As seen for instance in Figure~\ref{fig:compare-losses-and-alphas}, a combination of a restrictive loss ($\tcov = 1.0)$ and a small $\alpha=0.01$ would entail selecting (almost) all classes for every pixel.
This can be taken as a diagnostic signal: the model is not good enough for our notion of risk, we need to either augment our tolerance for errors or revise the prediction model, for instance.

\section{Results}
\label{sec:results}

\begin{table}
    \centering
    \begin{tabular}{l l l cr}
          \textbf{Dataset} 
            & \multicolumn{1}{l}{$\alpha$} 
            & \multicolumn{1}{c}{$\tcov$}
            & \multicolumn{1}{c}{\textbf{Empirical Risk}}
            & \multicolumn{1}{c}{\textbf{AR}} \\
        \midrule
        Cityscapes
         & 0.1 & 0.99 & 0.106  {\scriptsize $\pm$ (0.019)} & 1.028 \\
         & 0.1 & 0.95 & 0.100  {\scriptsize $\pm$ (0.021)} & 1.274 \\
         & 0.01 & 0.95 & 0.011 {\scriptsize $\pm$ (0.008)} & 1.208 \\
         & 0.01 & 0.99 & 0.011 {\scriptsize $\pm$ (0.014)} & 2.557 \\
         & \vspace{-0.5em} \\

        ADE20K         
         & 0.1  & 0.75  & 0.082 {\scriptsize ($\pm$ 0.021)} & 1.440 \\
         & 0.1  & 0.90  & 0.076 {\scriptsize ($\pm$ 0.021)} & 3.483 \\
         & 0.01 & 0.75  & 0.004 {\scriptsize ($\pm$ 0.005)} & 9.349 \\
        & \vspace{-0.5em}\\
        LoveDA
         & 0.10 & 0.50 & 0.097 {\scriptsize ($\pm$ 0.018)} & 1.231 \\
         & 0.10 & 0.75 & 0.103 {\scriptsize ($\pm$ 0.013)} & 2.672 \\
         & 0.10 & 0.90 & 0.092 {\scriptsize ($\pm$ 0.012)} & 3.946 \\
         & 0.01 & 0.50 & 0.010 {\scriptsize ($\pm$ 0.008)} & 3.607\\
         & 0.01 & 0.75 & 0.010 {\scriptsize ($\pm$ 0.005)} & 4.956\\
         & 0.01 & 0.90 & 0.010 {\scriptsize ($\pm$ 0.006)} & 5.761\\

    \end{tabular}
    \caption{{Metrics on $\Dtest$: empirical risk and activation ratio (AR) for \textbf{binary loss} $\ell_{\tcov}$. Empirical risk should be as close as possible to $\alpha$ to show validity.} 
    For each line, we repeat several times this procedure: (1) shuffle the dataset, (2) split validation data into $\Dcal$ \& $ \Dtest$, (3) run calibration on $\Dcal$, (4) Compute metrics on $\Dtest$. We finally average the metrics over the multiple runs (standard deviation in the parentheses).
    }
    \label{tab:metrics-binary}
\end{table}

\begin{table}
    \centering
    \begin{tabular}{l l c r}
        \textbf{Dataset} &  \multicolumn{1}{l}{$\alpha$ } & \multicolumn{1}{c}{\textbf{Empirical Risk}} & \multicolumn{1}{c}{\textbf{AR}} \\
        \midrule
        Cityscapes
         & \multicolumn{1}{l}{0.05}  & 0.041 {\scriptsize ($\pm$ 0.001)} & 1.000$^{\dagger}$ \\
         & \multicolumn{1}{l}{0.01}  & 0.006 {\scriptsize ($\pm$ 0.001)} & 1.230 \\
         & \multicolumn{1}{l}{0.005} & 0.001 {\scriptsize ($\pm$ 0.0003)}& 1.998 \\
         & \vspace{-0.5em} \\
        ADE20K
         & 0.2                      & 0.179 {\scriptsize ($\pm$ 0.005)} &   1.000$^{\dagger}$ \\
         & \multicolumn{1}{l}{0.1}  & 0.098 {\scriptsize ($\pm$ 0.008)} &   1.362 \\
         & \multicolumn{1}{l}{0.05} & 0.048 {\scriptsize ($\pm$ 0.006)} &   2.474 \\
         & \multicolumn{1}{l}{0.01} & 0.008 {\scriptsize ($\pm$ 0.002)} &  15.285 \\
                       & \vspace{-0.5em}\\ 
        LoveDA
         & \multicolumn{1}{l}{0.2}   & 0.199 {\scriptsize $\pm$ (0.009)}  & 1.388\\
         & \multicolumn{1}{l}{0.1}   & 0.100 {\scriptsize $\pm$ (0.006)}  & 2.650\\
         & \multicolumn{1}{l}{0.05}  & 0.049 {\scriptsize $\pm$ (0.004)}  & 4.069\\
         & \multicolumn{1}{l}{0.01}  & 0.008 {\scriptsize $\pm$ (0.0008)} & 6.350\\
         & \multicolumn{1}{l}{0.005} & 0.003 {\scriptsize $\pm$ (0.0004)} & 6.796\\

    \end{tabular}
    \caption{{Metrics on $\Dtest$: empirical risk and activation ratio (AR) for \textbf{miscoverage loss} $\ell$. 
    $^{\dagger}$: the underlying predictor attains the risk level without adding any class to the prediction set, that is, the output semantic mask with one class per pixel already satisfies this risk level.
    }}
    \label{tab:metrics-miscoverage}
\end{table}

In \Cref{tab:metrics-binary} and \Cref{tab:metrics-miscoverage}, we report the results of our experiments.
For both the empirical risk and the activation ratio (AR), we average the metrics over multiple runs (ten) of each loss configuration.

As it is customary in \cp, one first ensures that the theoretical guarantees holds also in practice for a given dataset and predictive model.
As expected, the empirical risks are a very close approximation of the nominal risk values $\alpha$.
As for the activation ratios, they follow a reasonable pattern: they increase as the notion of error gets stricter.
In \cref{tab:metrics-miscoverage}, we see how the AR increases when the risk diminishes from $\alpha=0.05$ to $\alpha=0.01$.

In \cref{fig:intro-segmask-varisco} and \cref{fig:visu-conformal-maps} we give examples of conformalized predictions for the three datasets.
In all cases, the borders of the masks often appear to be highlighted as more uncertain.
For the other patches in darker shades of red, the signal of uncertainty produced by the conformalized predictors go towards more ambiguous areas, smaller and farther objects.

\section{Conclusion}
\label{sec:conclusion}
Thanks to its light computational footprint,
\cp can be applied wherever \ml is deployed in sensitive or critical applications, with limited knowledge of the underlying model (e.g. black box).
We have shown how to extend the theoretical guarantees of \cp to any predictor for semantic image segmentation, using a post-hoc a procedure that only requires access to softmax scores.

In the future, it could be interesting to study the interactions of our method with existing \uq predictors (\eg bayesian).
Another promising direction would also be to extend this work to panoptic segmentation, which would allow to extend the theoretical guarantees to instances in the input images.
From the point of view of safety in AI, one could also profit from experimenting with class-conditional conformal guarantees: one can restrict \crc to a subset of priority classes, such as pedestrians or bicycle riders in automonous driving, ignoring the others.
Finally, a major methodological challenge would be to work towards providing theoretical guarantees to sequences of data, notably for real-time video.

\section*{Acknowledgments}
This work has benefited from the support of the DEEL project,\footnote{\url{https://www.deel.ai/}} with fundings from the Agence Nationale de la Recherche, and which is part of the ANITI AI cluster.

{
    \small
    \bibliographystyle{ieeenat_fullname}
    \bibliography{main}
}

\end{document}